# Integrated Decision and Control at Multi-Lane Intersections with Mixed Traffic Flow

Jianhua Jiang[1, 2]
Tsinghua University
Beijing, China
jiangjianhua_1998@163.com

Yangang Ren[1]
Tsinghua University
Beijing, China
ryg18@mails.tsinghua.edu.cn

Yang Guan[1]
Tsinghua University
Beijing, China
guany17@mails.tsinghua.edu.cn

Shengbo Eben Li*[1]
Tsinghua University
Beijing, China
lisb04@gmail.com

Yuming Yin[1]
Tsinghua University
Beijing, China
yinyuming89@gmail.com

Dongjie Yu[1]
Tsinghua University
Beijing, China
ydj20@mails.tsinghua.edu.cn

Xiaoping Jin[2]
China Agricultural University
Beijing, China
jinxp@cau.edu.cn

## ABSTRACT
Autonomous driving at intersections is one of the most complicated and accident-prone traffic scenarios, especially with mixed traffic participants such as vehicles, bicycles and pedestrians. The driving policy should make safe decisions to handle the dynamic traffic conditions and meet the requirements of on-board computation. However, most of the current researches focuses on simplified intersections considering only the surrounding vehicles and idealized traffic lights. This paper improves the integrated decision and control framework and develops a learning-based algorithm to deal with complex intersections with mixed traffic flows, which can not only take account of realistic characteristics of traffic lights, but also learn a safe policy under different safety constraints. We first consider different velocity models for green and red lights in the training process and use a finite state machine to handle different modes of light transformation. Then we design different types of distance constraints for vehicles, traffic lights, pedestrians, bicycles respectively and formulate the constrained optimal control problems (OCPs) to be optimized. Finally, reinforcement learning (RL) with value and policy networks is adopted to solve the series of OCPs. In order to verify the safety and efficiency of the proposed method, we design a multi-lane intersection with the existence of large-scale mixed traffic participants and set practical traffic light phases. The simulation results indicate that the trained decision and control policy can well balance safety and tracking performance. Compared with model predictive control (MPC), the computational time is three orders of magnitude lower.

## CCS Concepts
• **Theory of computation**
~ Theory and algorithms for application domains
~ Machine learning theory
~ Reinforcement learning
~ Sequential decision making

*This work is supported by NSF China with U20A20334 and 51575293. It is also partially supported by Tsinghua University-Toyota Joint Research Center for AI Technology of Automated Vehicle. J. Jiang and Y. Ren contributed equally to this work.*

*[1]School of Vehicle and Mobility, Tsinghua University, Beijing, 100084, China. [2]College of Engineering, China Agricultural University, Beijing, 100083, China. All correspondence should be sent to Shengbo Eben Li. <lisb04@gmail.com>.*

## 1. INTRODUCTION
Autonomous driving is a potential technology to alleviate traffic jams, enhance driving safety and allow total freedom of movement. As the typical scenario of urban transportation, the intersection is suitable and essential to verify and apply related algorithms, but it is challenging due to the complicated environment with random, dynamic and mixed traffic flow. According to the official statistics published by the USA, fatal automotive crashes related to intersections account for nearly 30% of all car accidents, which causes 18% of pedestrian fatalities [1]. Moreover, the mapping relationship of mixed traffic participants at the intersections requires improving computational efficiency to avoid collisions. Hence, developing safe decision and control strategies with fast response performance at the complex intersections with mixed traffic flow is significant for popularizing autonomous driving.

At present, most of the solutions for autonomous driving can be categorized into two major frameworks: hierarchical decision and control and integrated decision and control. The first framework decomposes the whole process into independent subtasks [2], such as scenario understanding, motion prediction and trajectory planning, while each subtask is solved separately to achieve final goals [3]-[4]. This framework has been investigated and applied in the industry to deal with some scenes during these years. Glaser *et al.* presented an algorithm for vehicle path planning adapted to traffic on a lane-structured infrastructure such as highways [5]. Kala et al. proposed a planning algorithm for behavior selection based on obstacle movement by modeling the various maneuvers in an empirical formula [6]. Bojarski *et al.* built an end-to-end system to perform lane keeping in some scenarios such as highways and local roads, which needs mass data to train [7]. However, under the hierarchical decision and control framework, most of the subtasks require longer computational time as the complexity of traffic scenes increases [8]-[9], especially for scenarios with mixed traffic participants.

Another one integrates the decision and control tasks into a unified framework, consisting of two modules called static path planning and optimal dynamic tracking. The former is used to generate multiple paths only considering static constraints like road structure, and the latter aims to select and track the path considering dynamic traffic participants and traffic lights. A constrained OCP is formulated for each static path, which can optimize the different state constraints during driving. Guan *et al.* first proposed the integrated decision and control framework and solved the OCP offline by a model-based RL algorithm called generalized exterior point (GEP) to seek an approximate solution of an OCP in the form of neural networks. They proved this framework's superiority through simulations and real road tests compared to the hierarchical decision and control [10]-[11]. However, their research only verified the algorithm into scenarios with a single type of traffic participant, which cannot reflect the hybrid characteristics of traffic flow in the real world. Besides, this framework cannot identify the information of traffic lights because the traffic light state is not considered as the state input.

This paper focuses on the multi-lane intersections with mixed traffic flow and proposes a learning-based decision and control algorithm for this scenario, which can deal with mixed traffic participants and has high online computational efficiency. The main contributions of this paper are summarized as follows:

1) We utilize and improve the integrated decision and control framework to deal with the complex mixed traffic flow at the intersection and use RL to solve this problem, extending the simulation to the more complicated and realistic scenario and reducing computation overhead significantly.

2) Take the traffic light state as the system input, the improved framework can recognize the process the different traffic signals. The states and constraints are also redesigned to deal with the mixed traffic flow. Besides, we plan the changing expected velocity curve, and first proposed a finite state machine to select the curve reasonably.

This paper is organized as follows. Section 2 describes the improved integrated decision and control framework for mixed traffic flow. Section 3 illustrates the details of the problem statement and methodology. Section 4 looks into simulated settings and results illustrations, and last section 5 summarizes this work.

## 2. FRAMEWORK

This section mainly introduces the improved integrated decision and control framework at multi-lane intersections with mixed traffic flow. The process of applying the framework to this scenario is illustrated in Fig. 1, which consists of two modules: static path planning and optimal path tracking.

The first module aims to generate several candidate paths only considering static information such as road structure and traffic lights. The set of static paths will be designed to include position information and expected velocity according to the traffic light states. As shown in the top half of Fig. 1, the curves of expected stop and pass velocity are different, and the numerical value is piecewise functions that take the entrance and exit of the intersection as the dividing points. When the traffic light is green, the expected pass velocity is selected, which velocity inside the intersection is slightly lower than that outside the intersection area. In comparison, the expected stop velocity at a red light will gradually decrease to 0. As for the expected velocity at the yellow light, it can be expressed as one of the above two by a finite vector machine. Note that these paths are not optimal trajectories but basis references for optimization. Compared to the traditional schemes, this module is more efficient because the shape and generation method of the static path is not relatively complex, and it has high embeddability and scalability to apply into the electronic map as long as the key parameters of the road are obtained.

Another module called optimal path tracking is the core segment of this framework, which further considers the candidate paths, traffic lights and traffic participants around the intersection. To ensure collision avoidance in the mixed traffic, this module establishes safety constraints for the ego vehicle and specific traffic participants, especially the pedestrians and bicycles with potential conflicts. The states and constraints are redesigned in this improved framework. A constrained OCP is formulated for each static path generated by the first module, which minimizes the tracking error and constraints characterized by safety and efficiency requirements within a finite horizon. And then, the path with the least cost will be selected as optimal reference to track in each step. Specially, we use a model-based RL algorithm to solve these OCPs in the form of two neural networks (NNs). This module can settle this problem of low computational efficiency in the online optimization process due to the fast propagation of NNs. Therefore, the improved integrated decision and control framework is the potential to apply to automotive on-board controllers to achieve real-time calculation.

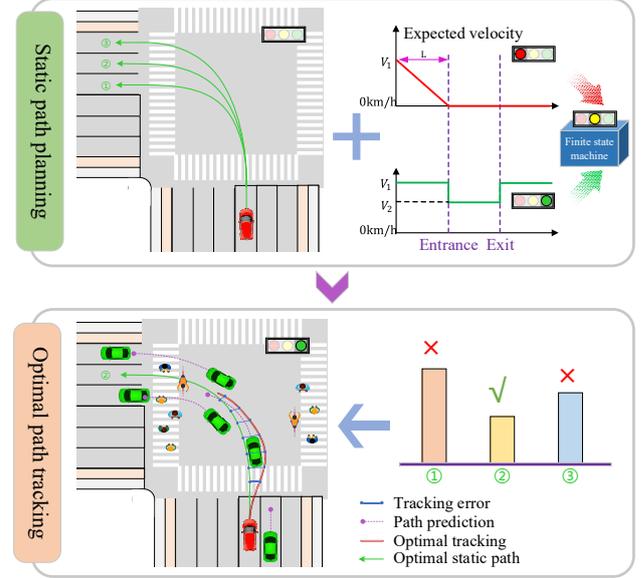

**Figure 1.** The improved integrated decision and control framework at intersections with mixed traffic flow

## 3. METHODOLOGY

Here, we mainly focus on the improvements of integrated decision and control framework for the intersection with mixed traffic flow, including the reconstruction of the OCP and the design of states and constraints. Besides, the method of solving the constrained OCP using RL will also be involved.

### 3.1 Static path planning

The traffic light system has three states, i.e., green, yellow and red lights, and we plan two kinds of expected velocity curves shown as Fig. 1. In response to the traffic light state, the improved version of static path planning is given, as shown in Fig. 2, which utilizes a finite state machine to deal with the different traffic signals. In Fig. 2, A represents current traffic light states, including green, yellow and red lights. Condition B indicates if A is the yellow light, whether the ego vehicle can stop in front of the stop line at the current position at a deceleration that does not affect the comfort. When the ego vehicle enters the intersection area, the expected velocity is selected according to the current signal light state. Intuitively, at the red light or green light, the ego vehicle just needs to select the corresponding speed curve as the expected velocity. As a warning signal, the yellow light allows the vehicle to choose to wait or drive according to the traffic in the intersection and vehicle states. If the ego vehicle observes that it is difficult to stop in the designated area and the traffic in the intersection is not crowded, the expected pass velocity is selected. Otherwise, another one will be chosen. Hence, the proposed finite state machine model can choose a well expected velocity curve under different traffic lights and further ensure safe driving.

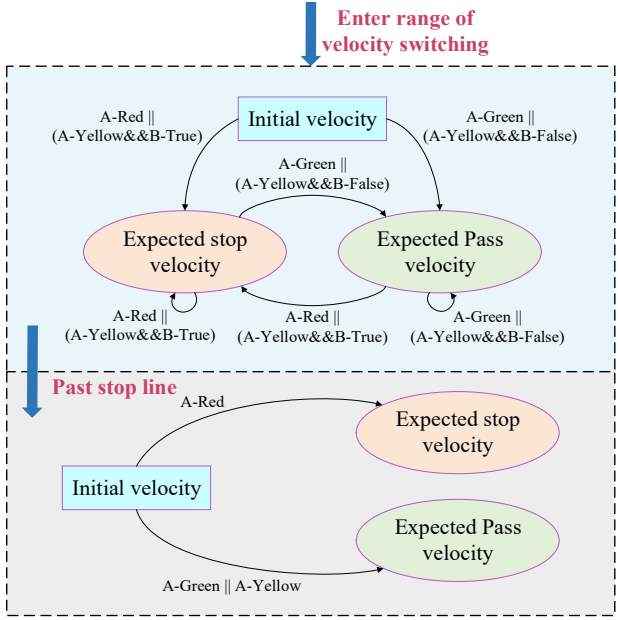

**Figure 2.** Finite state machine for expected velocity switching

## 3.2 Optimal path tracking

This module is reconstructed because of the mix of traffic participants and the traffic light state, which is reflected in the redesign of states and constraints and the reconstruction of the constrained OCP.

### 3.2.1 Consideration of states

When the ego vehicle conducts different tasks, it needs to consider different surrounding traffic participants at the intersection, so we encode a certain number of vehicles, bicycles or pedestrians with potential conflicts according to different tasks. Take the left turn task as an example. Vehicles and bicycles are encoded by their respect route start and end, as well as the order on that, while only pedestrians walking at the left crosswalk are considered. Similarly, the straight driving and right turn tasks also can be defined in the same way. The state is designed to include information of the ego vehicle, surrounding traffic participants, static paths and traffic light state. Fig. 3 shows the surrounding traffic participants considered by ego vehicle in three tasks at the scenario, in which the ego vehicle is blue, and the shapes with different color represent routes of traffic participants considered by the ego vehicle. Besides, the traffic light state is also considered as the input, whose phases are represented by a number.

### 3.2.2 Design of constraints

Due to the addition of pedestrians and bicycles, the number of constraints between the ego vehicle and surrounding participants increases greatly, so it is necessary to simplify these constraints. Since the appearance of the traffic participants is almost rectangular, their shape should be considered to calculate the minimum distance. Based on the isotropy of the circle, we utilize double circles to approximate the shape of vehicles and bicycles, and the single circle is used to represent pedestrians, as illustrated in Fig. 4. Finally, there are four constraints between the ego vehicle and each vehicle or bicycle around, and two constraints between the ego vehicle and pedestrian.

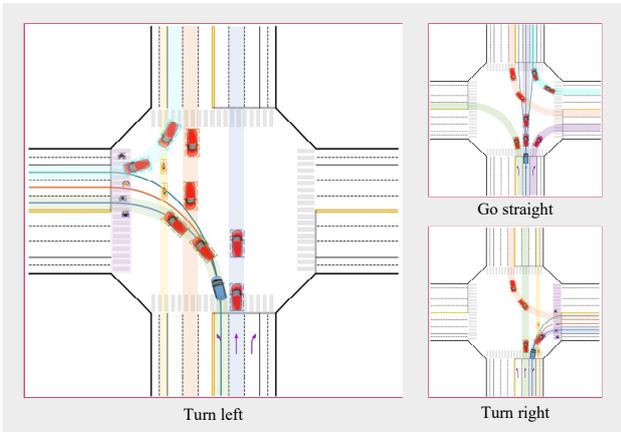

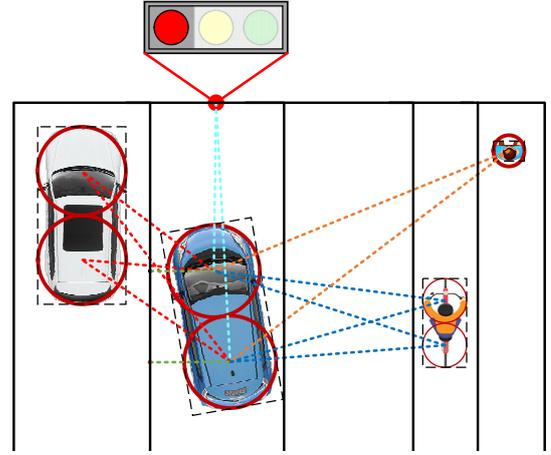

**Figure 3.** Surrounding participants considered by the ego vehicle

**Figure 4.** Constraint design and distance representation

Instead of using virtual vehicles to simulate red light signals, we set constraints at the center of the stop line, which only works when the red light is on. The distance between the ego vehicle and the road can be expressed by the distance between the center point of the circle and the edge line of the road. Therefore, we can define the safe distance as follows.

$$\begin{cases} d_{safe\_veh} = r_{ego} + r_{veh} \\ d_{safe\_bike} = r_{ego} + r_{bike} \\ d_{safe\_ped} = r_{ego} + r_{ped} \\ d_{safe\_road} = r_{ego} \\ d_{safe\_red} = r_{ego} \end{cases} \quad (1)$$

where $d_{(\cdot)}$ represents the minimum distance to be kept from the surrounding traffic participants, road edge and stop line to ensure safety, and $r_{(\cdot)}$ denotes the radius of the circle that characterizes ego vehicle and surrounding vehicles, bicycles and pedestrians.

### 3.2.3 Reconstructed of the constrained OCP

Given the static paths generated by the first module, we have extended the finite horizon constrained OCP for each path in [11] to (2). In this problem, the tracking error will be minimized, and the safety constraints will also be satisfied.

$$\min_{u_{i|t}} \quad J = \sum_{i=0}^{T-1} (x_{i|t}^{ref} - x_{i|t})^\top Q(x_{i|t}^{ref} - x_{i|t}) + u_{i|t}^\top R u_{i|t}, i = 0:T-1 \quad (2)$$
$$s.t. \quad x_{i+1|t} = f_{ego}(x_{i|t}, u_{i|t}),$$

$$x_{i+1|t}^{j\_veh} = f_{sur\_veh}(x_{i|t}^{j\_veh}),$$
$$x_{i+1|t}^{j\_bike} = f_{sur\_bike}(x_{i|t}^{j\_bike}),$$
$$x_{i+1|t}^{j\_ped} = f_{sur\_ped}(x_{i|t}^{j\_ped}),$$
$$g(x_{i|t}) = \begin{cases} (x_{i|t} - x_{i|t}^{j\_veh})^\top W(x_{i|t} - x_{i|t}^{j\_veh}) - d_{safe\_veh} \\ (x_{i|t} - x_{i|t}^{j\_bike})^\top W(x_{i|t} - x_{i|t}^{j\_bike}) - d_{safe\_bike} \\ (x_{i|t} - x_{i|t}^{j\_ped})^\top W(x_{i|t} - x_{i|t}^{j\_ped}) - d_{safe\_ped} \\ (x_{i|t} - x_{i|t}^{road})^\top W(x_{i|t} - x_{i|t}^{road}) - d_{safe\_road} \\ (x_{i|t} - x_{i|t}^{red})^\top W(x_{i|t} - x_{i|t}^{red}) - d_{safe\_red} \end{cases} > 0, \quad (2)$$
$$x_{0|t} = x_t, u_{0|t} = u_t,$$
$$x_{0|t}^{j\_veh} = x_t^{j\_veh}, x_{0|t}^{j\_bike} = x_t^{j\_bike}, x_{0|t}^{j\_ped} = x_t^{j\_ped}$$

where $t$ is the current time step, $i$ and $T$ are the virtual time step and predictive horizon, respectively. Note that virtual time means time in finite horizon. $x_{i|t}$ and $u_{i|t}$ represent the state and control action of the ego vehicle at the virtual time $i$, $x_{i|t}^{ref}$ is the closest reference point from ego vehicle to the static path. $f_{ego}$ represents the dynamic model of ego vehicle, while $f_{sur\_veh}$, $f_{sur\_bike}$ and $f_{sur\_ped}$ are the kinematics model of surrounding vehicle, bike and pedestrian. Besides, $x_{i|t}^{j\_veh}$, $x_{i|t}^{j\_bike}$ and $x_{i|t}^{j\_ped}$ are the state of traffic participants, in which $j$ denote the order number of them. $x_{i|t}^{road}$ is the state of road edge that closest to the ego vehicle, and $x_{i|t}^{red}$ define the position of stop line when traffic light is red. $g(x_{i|t})$ denotes the safety constraints between ego vehicle and traffic participants, road edge and stop line, and $W$ is defined as the weighted matrix.

In order to simplify this problem, the constraint can be added to the objective function in the form of a penalty function, which includes the penalty for each constraint. Hence, the objective function in (2) can be rebuilt to (3), and the original problem becomes an unconstrained control problem.

$$\min_{u_{i|t}} \quad J = \sum_{i=0}^{T-1} l(x_{i|t}, u_{i|t}) + \rho\varphi(x_{i|t}), i = 0: T-1$$
$$l(x_{i|t}, u_{i|t}) = (x_{i|t}^{ref} - x_{i|t})^\top Q(x_{i|t}^{ref} - x_{i|t}) + u_{i|t}^\top R u_{i|t} \quad (3)$$
$$\varphi(x_{i|t}) = [max\{0, -g(s_{i|t})\}]^2$$

where $\varphi(x_{i|t})$ is the penalty function, and $\rho$ is the penalty factor.

To balance the facticity and efficiency of the calculation, we adopt the 3-Degree-of-Freedom (DOF) single-track model for the ego vehicle, as shown in (4). This model is discretized utilized the backward Euler method, which guarantees its numerical stability at any low speed [12].

$$f_{ego} = \begin{bmatrix} x + \Delta t(v_{lon}\cos\varphi - v_{lat}\sin\varphi) \\ y + \Delta t(v_{lon}\sin\varphi + v_{lat}\cos\varphi) \\ v_{lon} + \Delta t a \\ \dfrac{m v_{lon} v_{lat} + \Delta t[(L_f k_f - L_r k_r)\omega - k_f \delta v_{lon} - m v_{lon}^2 \omega]}{m v_{lon} - \Delta t(k_f + k_r)} \\ \varphi + \Delta t\omega \\ \dfrac{-I_z \omega v_{lon} - \Delta t[(L_f k_f - L_r k_r)v_{lat} - L_f k_f \delta v_{lon}]}{\Delta t(L_f^2 k_f + L_r^2 k_r) - I_z v_{lon}} \end{bmatrix} \quad (4)$$

where $\Delta t$ is the control step of actuator, $[x, y, v_{lon}, v_{lat}, \varphi, \omega]^\top$ denotes the state of ego vehicle, which includes x and y coordinates, longitudinal and lateral velocities, heading angle and yaw rate. $[\delta, a]^\top$ is the control vector that outputs the front wheel angle and the acceleration commands. Besides, $m$ and $I_z$ are the weight and moment of inertia, respectively. $L_f$ and $L_r$ represent the distance from front/rear axle center to vehicle centroid, and $k_f$ and $k_r$ mean the front/rear side slip stiffness.

The state of surrounding traffic participants is represented as $[x_{sur}, y_{sur}, v_{sur\_lon}, \varphi_{sur}]^\top$, and the model is indicated as (5).

$$f_{sur} = \begin{bmatrix} x_{sur} + \Delta t(v_{sur\_lon}\cos\varphi_{sur}) \\ y_{sur} + \Delta t(v_{sur\_lon}\sin\varphi_{sur}) \\ v_{sur\_lon} \\ \varphi_{sur} + \Delta t\omega_{sur} \end{bmatrix} \quad (5)$$

Because the constrained OCP is formulated for a single path, and there are multiple static paths generated by the first module, the optimal path is selected in terms of a rule-based safety indicator, as shown below.

$$p^* = \underset{p}{\operatorname{argmin}}\{J_1, J_2, J_3\} \quad (6)$$

### 3.2.4 Solving with RL

This constrained OCP is constructed for each static path, which can be solved by traditional online optimization methods such as MPC or offline optimization methods like RL. The online algorithm has little efficiency in the calculation process, so it is difficult to meet the requirements of real-time computing. In comparison, the RL method can find a neural network solution for the constrained OCP with offline training to obtain high computational efficiency. However, there are certain differences between the above OCP and the RL problems, so that this offline method cannot be applied directly. We rebuild a complete RL problem formulation based on the OCP (2), and it is represented in (7) and (8), which consists of the actor that maps from state to action and the critic that evaluates action performed. Besides, this RL problem does not seek to find a single optimal control quantity of a single state based on the specific static path but to solve the optimal parameters of actor and critic with a state distribution and multiple static paths. The state is designed to contain certain necessary information, including the information of ego vehicle, surrounding traffic participants, static paths and traffic signals.

$$\min_\theta \quad J_p = J_{actor} + \rho J_{penalty}$$
$$= \mathbb{E}_{s_{0|t}}\left\{\sum_{i=0}^{T-1} l(s_{i|t}, \pi_\theta(s_{i|t}))\right\} + \rho \mathbb{E}_{s_{0|t}}\left\{\sum_{i=0}^{T-1} \varphi_i(\theta)\right\},$$
$$i = 0: T-1$$
$$l(s_{i|t}, \pi_\theta(s_{i|t})) = (x_{i|t}^{ref} - x_{i|t})^\top Q(x_{i|t}^{ref} - x_{i|t}) + \pi_\theta(s_{i|t})^\top R \pi_\theta(s_{i|t})$$
$$\varphi_i(\theta) = [max\{0, -g(s_{i|t})\}]^2$$
$$s.t. \quad s_{i+1|t} = f(s_{i|t}, \pi_\theta(s_{i|t})), \quad (7)$$
$$g(s_{i|t}) = \begin{cases} (x_{i|t} - x_{i|t}^{j\_veh})^\top W(x_{i|t} - x_{i|t}^{j\_veh}) - d_{safe\_veh} \\ (x_{i|t} - x_{i|t}^{j\_bike})^\top W(x_{i|t} - x_{i|t}^{j\_bike}) - d_{safe\_bike} \\ (x_{i|t} - x_{i|t}^{j\_ped})^\top W(x_{i|t} - x_{i|t}^{j\_ped}) - d_{safe\_ped} \\ (x_{i|t} - x_{i|t}^{road})^\top W(x_{i|t} - x_{i|t}^{road}) - d_{safe\_road} \\ (x_{i|t} - x_{i|t}^{red})^\top W(x_{i|t} - x_{i|t}^{red}) - d_{safe\_red} \end{cases} > 0,$$
$$s_{i|t} = [p^{ref}, x_{i|t}, x_{i|t}^{j\_veh}, x_{i|t}^{j\_bike}, x_{i|t}^{j\_ped}, x_{i|t}^{phase}]^\top$$

$$\min_\omega \quad J_{critic} = \mathbb{E}_{s_{0|t}}\left\{\left(\sum_{i=0}^{T-1} l(s_{i|t}, \pi_\theta(s_{i|t})) - V_\omega(s_t)\right)^2\right\}, i = 0: T-1$$
$$s.t. \quad s_{i+1|t} = f(s_{i|t}, \pi_\theta(s_{i|t})), \quad (8)$$
$$s_{i|t} = [p^{ref}, s_{i|t}, s_{i|t}^{j\_veh}, s_{i|t}^{j\_bike}, s_{i|t}^{j\_ped}, s_{i|t}^{phase}]^\top$$

where $s_{0|t} = s_t \leftarrow \{p^{ref}, s_{i|t}, s_{i|t}^{j\_veh}, s_{i|t}^{j\_bike}, s_{i|t}^{j\_ped}, s_{i|t}^{phase}\}$ denotes the state includes information of static paths, ego vehicle state, surrounding traffic participants states and traffic light state, and $\pi_\theta(s_{i|t})$ is the parameterized policy called actor that represents the mapping from state to action. $f$ is the synthetization of $f_{ego}$, $f_{sur\_veh}$, $f_{sur\_bike}$ and $f_{sur\_ped}$. In the state, $p^{ref}$ represents the whole information of

static paths, and $x_{i|t}^{phase}$ is the traffic light state, and $g(s_{i|t})$ denotes the constraints about the state. $V_\omega(s_t)$ is the critic parameterized by $\omega$ in the form of NNs that similar to actor, and it satisfies the following definition of the optimality.

$$J^* = V_{\omega^*}(s_t) \tag{9}$$

After that, we can update the policy parameters and the penalty factor by gradient descent methods and given monotonic increasing sequence, respectively. Given the optimal policy after offline training, we can apply it and select the optimal static path based on a rule as shown in (10).

$$V(p) = \sum d(p, x^j) \tag{10}$$

where $d(p, x^{sur})$ represents the distance from $j$-th surrounding traffic participants to the path. Therefore, the optimal path can be select with the maximum value of $V$, and the optimal action $u_t^*$ can be obtained by the policy. However, this policy cannot guarantee the safety for each action due to incomplete sampling, which can be solved by utilizing safety shield. This method aims to add a safety shield before conduct the action and project the unsafe actions to the nearest safe action.

$$u_t^{safe} = \begin{cases} u_t^*, & if\ u_t^* \in \mathcal{U}_{safe}(s_t) \\ \arg\min_{u \in \mathcal{U}_{safe}(s_t)} \|u - u_t^*\|^2, & else \end{cases} \tag{11}$$

where $\mathcal{U}_{safe}(s_t)$ denotes the safe action space.

## 4. SIMULATION

### 4.1 Scenario and task description
The complexity and challenge of the scenario are shown in the addition of bicycle lanes, sidewalks and crosswalks at the regular signalized four-way intersection, and each lane or area has access rights for different types of traffic participants. There are three roads for motor vehicles, one road for bicycles and one sidewalk in each driving direction, and the typical intersection system, as shown in Fig. 5. Besides, we generate mixed traffic flow with high density based on SUMO [13], which passes through this scenario orderly according to the traffic light phases. A kind of traffic light time-distributing with more conflicts is adopted, which can better verify the ability of the algorithm to deal with complex scenarios. In particular, we do not separate the left turn from the straight driving to create more complicated intersection and allow the right turn all the time. The ego vehicle applied the algorithm is required to complete three tasks: turn left, go straight and turn right, and avoid collision with traffic participants or road edge.

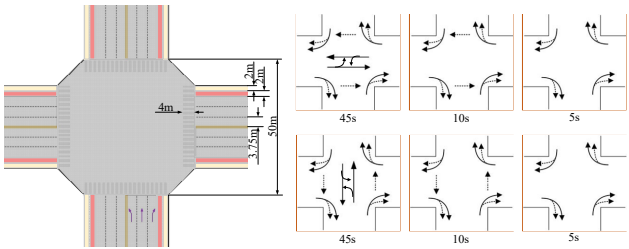

**Figure 5.** The intersection structure and traffic light phases

### 4.2 Static paths generation
The ideal static path should be smooth and located in the middle of the lane to ensure stable driving for the ego vehicle in the process of optimal path tracking, and the Cubic Bezier curve is adopted to generate the multiple static paths at intersections. Four control points needed for the Cubic Bezier curve are obtained through the road map, which positions are defined on the extension line of the straight road center to ensure the consistency of curvature at the straight road and the intersection. The visualization of static paths is shown in Fig. 6.

$$\begin{cases} x_{ref}(\tau) = x_0(1-\tau)^3 + 3x_1\tau(1-\tau)^2 + 3x_2\tau^2(1-\tau) + x_3\tau^3 \\ y_{ref}(\tau) = y_0(1-\tau)^3 + 3y_1\tau(1-\tau)^2 + 3y_2\tau^2(1-\tau) + y_3\tau^3 \\ \varphi_{ref}(\tau) = \arctan\left(\frac{y(\tau+1) - y(\tau)}{x(\tau+1) - x(\tau)}\right) \\ v_{ref} = v_{exp} \end{cases} \tag{12}$$

where $\tau \in [0,1]$ denotes the parameter of Cubic Bezier curve, $(x_0, y_0)$, $(x_1, y_1)$, $(x_2, y_2)$ and $(x_3, y_3)$ are the control points and shown as the solid points in Fig. 6, $v_{ref}$ is the expected velocity defined in Fig.1, $(x_{ref}, y_{ref}, \varphi_{ref}, v_{ref})$ is the output of the static path planning.

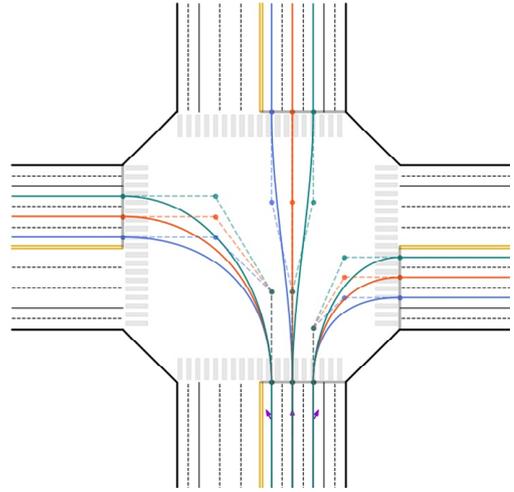

**Figure 6.** The multiple static paths generation at intersection

### 4.3 Training results
We utilize a multi-layer perception (MLP) to approximate value function and policy, which has two hidden layers that employ exponential linear units (ELU) as the activation function and contain 256 units per layer. The predictive horizon is set to be 25 with a frequency of 10Hz, and the curves of three tasks during training are shown in Fig. 7. Results show that the loss of each item has a significant downward trend and converges to a lower value quickly, which means the policy can satisfy each constraint defined after iteration converge. In order to validate the accuracy and feasibility of our method, the simulations in the above scenario with mixed traffic flow are given. The high-density traffic flow is generated with 800 vehicles, 100 bicycles and 450 pedestrians per hour at each lane, which composes a complicated scenario that can verify the algorithm effectively. Then we conduct the simulations for three tasks at the intersection with randomly generated traffic flow, respectively, and the control performance of our method for each task is shown in Fig. 8. The trained policy can realize a movement close to human driving, which is reflected in avoiding the surrounding traffic participants and passing behind bicycles and pedestrians at the intersection. Specifically, in the left-turn task, the ego vehicle avoids one vehicle coming from the opposite direction and passes quickly from the back of that then. When meeting bicycles and pedestrians, the ego vehicle will take a brake and wait for them to pass through the safe route to achieve this task. Similarly, the performances of straight driving and right turn also show that the intelligence and reliability of our algorithm, which can complete the normal traffic tasks at the complicated intersections with mixed traffic flow and achieve driving behaviors similar to human drivers.

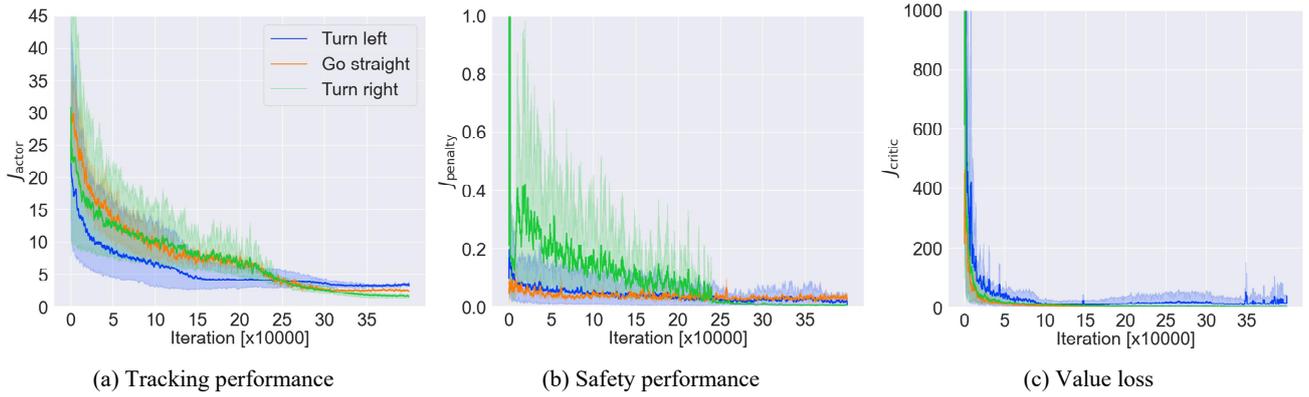

(a) Tracking performance   (b) Safety performance   (c) Value loss

**Figure 7.** Training results of the policies for three tasks

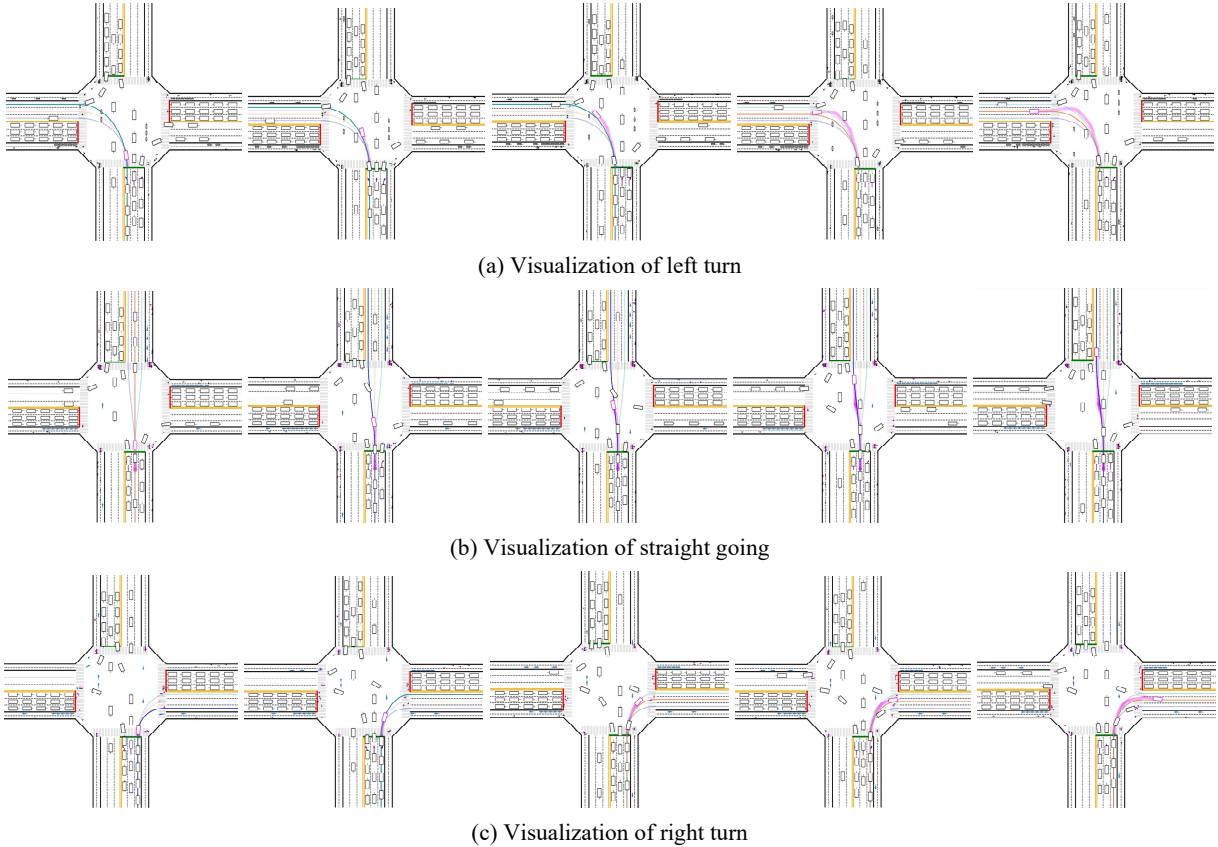

(a) Visualization of left turn

(b) Visualization of straight going

(c) Visualization of right turn

**Figure 8.** Visualization of three tasks using integrated decision and control framework

### 4.4 Switching of the expected velocity

Moreover, the simulations for left turn under different traffic light states are also carried out, and the curves of actual velocity and expected velocity are shown in Fig. 9. The solid line represents the actual velocity during the task, and the dotted line is the expected velocity that changes with the rule. Fig. 9a and 9b display the velocity curves at the green light and red light, which shows the expected velocity can be selected correctly with the light switching. Note that the actual velocity will not change the same as the expected velocity because of the limits of safety constraints in the driving process. Especially, the choice of velocity curves at yellow light is embodied in Fig. 9c and 9d, and the selection finite state machine model has been explained in section 3.1. When the ego vehicle cannot brake and stop in front of the stop line, the expected stop velocity is selected, as shown in Fig. 9c. Otherwise, the ego vehicle chooses another curve as the reference velocity.

### 4.5 Comparison with MPC

To verify the computational efficiency of the method proposed, we also contrast the algorithm with MPC in the same scenario and state. MPC uses the receding horizon optimization online and can obtain the exact solution for the constraints, which can be employed as a comparative method to test the accuracy of the results and compare the computational time. Besides, the Ipopt solver is adopted, an open-source package to solve large-scale nonlinear optimization problems using the Newton-type method [14]. We compare the two methods by defining the same states for the constrained OCP in the task of the right turn, and the actions solved are illustrated in Fig. 10a and 10b, and the results show the front wheel angle and the acceleration have similar trends with little errors. We also compare the computational time of the two methods in Fig. 10c. The results indicate that the time of our method to outputs actions is within 10ms, while MPC takes about 1000ms to calculate and output the optimal actions. Although the proposed method loses a small part of the accuracy, it can greatly reduce the computational time, which will be conducive to applying the algorithm to real vehicles and promoting the development of automated driving technology.

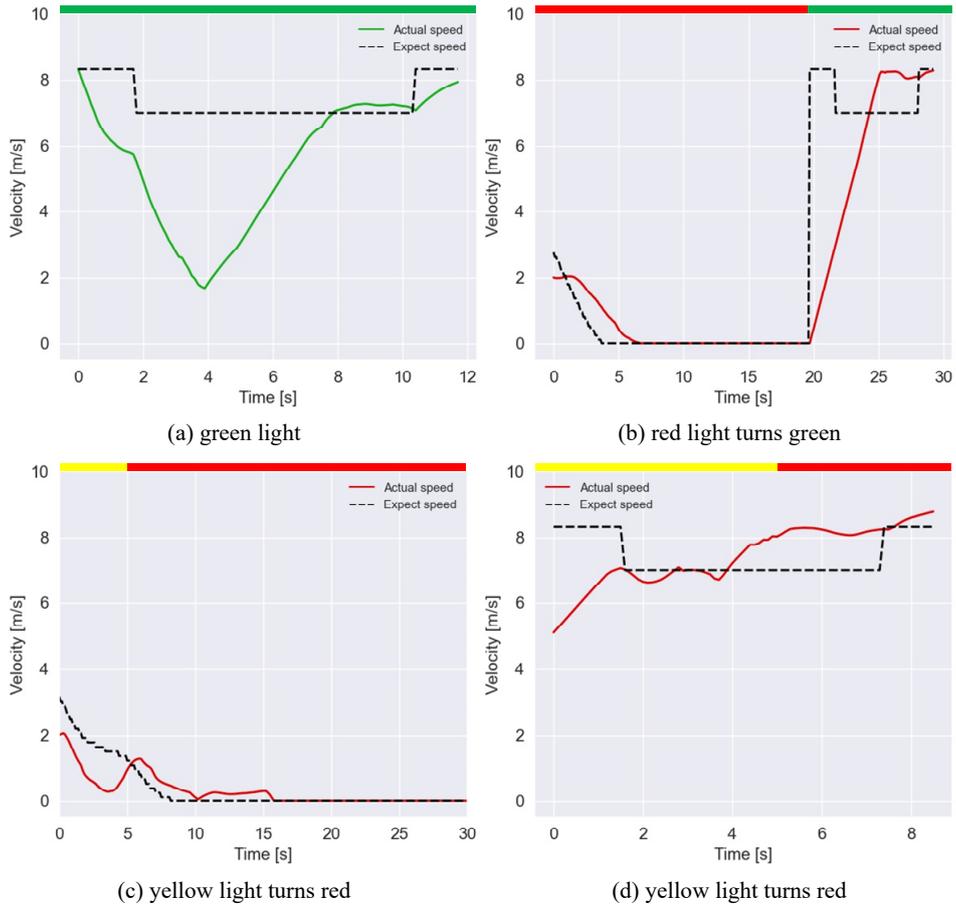

**Figure 9.** Velocity change with the different traffic lights

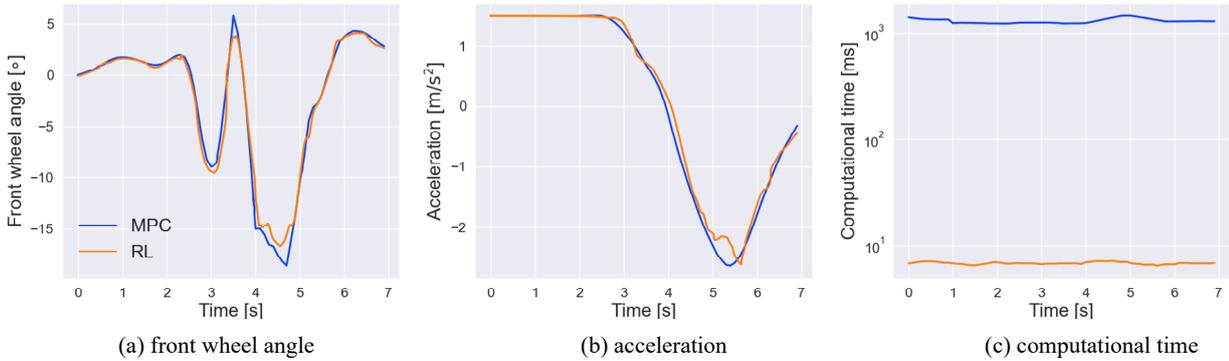

**Figure 10.** Comparison with MPC in right turn task

## 5. CONCLUSION

This paper focuses on the autonomous driving scenario at multi-lane intersections with mixed traffic flow. We implemented and improved the integrated decision and control framework, consisting of static path planning and optimal path tracking. In the static path planning, we define two kinds of the expected velocity curves and utilized a finite state machine to select a correct velocity curve under the different traffic signals. In the optimal path tracking, we reconstructed the constrained OCP for each static path, which adds new constraints between the ego vehicle and pedestrians, bicycles and the stop line at the red light. Meanwhile, the traffic light state is also utilized as input, and we design the representation of states and constraints for mixed traffic flow. Then, this problem is solved using model-based RL by offline training and online application. Moreover, we conduct the simulations in the intersection with mixed traffic flow, which demonstrates that it can realize safe decision-making under different tasks and changing traffic lights. Compared with the MPC, the computational time is reduced by approximately three orders of magnitude. In the future, we will handle dynamic number of surrounding participants and train one policy network to deal with different tasks.

## ACKNOWLEDGMENTS

This work is supported by NSF China with U20A20334 and 51575293. It is also partially supported by Tsinghua University-Toyota Joint Research Center for AI Technology of Automated Vehicle. We would also like to acknowledge Qi Sun and Jingliang Duan for their valuable suggestions for this paper.